\documentclass[lnbip]{svmultln}
\usepackage{subfigure}
\usepackage{graphicx}
\usepackage{makeidx}  
\begin{document}
\mainmatter              
\title{Satellite imagery analysis for operational damage assessment in Emergency situations}
\titlerunning{Damage assessment}  
%
\author{Alexey Trekin\inst{1} \and German Novikov\inst{1} \and Georgy Potapov\inst{1} \and Vladimir Ignatiev\inst{1} \and Evgeny Burnaev\inst{1}}
\authorrunning{Alexey Trekin et al.}   
%
\tocauthor{Alexey Trekin, German Novikov, Georgy Potapov, Vladimir Ignatiev, Evgeny Burnaev}
\institute{Skolkovo institute of Science and technology, Moscow, Nobel st. 1, Russia,\\
\email{aeronetlab@skoltech.ru},\\ WWW home page:
\texttt{http://crei.skoltech.ru/cdise/aeronet-lab/}}

\maketitle              

\begin{abstract}        
When major disaster occurs the questions are raised how to estimate the damage in time to support the decision making process and relief efforts by local authorities or humanitarian teams. 
In this paper we consider the use of Machine Learning and Computer Vision on remote sensing imagery to improve time efficiency of assessment of damaged buildings in disaster affected area.
We propose a general workflow that can be useful in various disaster management applications, and demonstrate the use of the proposed workflow for the assessment of the damage caused by the wildfires in California in 2017.

\end{abstract}
\section{Introduction}

In emergency situations like devastating wildfires, floods, earthquakes or tsunami, decision makers need to get information about possible damages in residential area and infrastructure most rapidly after the event. One of the most valuable sources to get such information from, is the Earth Observation systems, which include satellite and aerial remote sensing, since it can be captured shortly after the disaster without all the risks related to the ground observations. The combination of this information with statistical and ground observation data contributes to even a better valuation of physical and human losses caused by disaster \cite{eguchi}.

There are several international programs that are arranged to support the information exchange during and after disasters such as UNOSAT (UNITAR Operational Satellite Applications Program) \cite{unosat}, Space disaster charter \cite{sdc}, Humanitarian Openstreetmap Team (HOT) \cite{hot} or Tomnod \cite{tomnod} which is the Digitalglobe satellite company crowdsourcing platform. All these are useful initiatives providing tools and activation workflows for emergency mapping done by specialists or by volunteers in a collaborative way \cite{lang,dittus}.

This method of mapping of the imagery could be time-consuming since it requires some qualification and takes time to digitize all the damages manually, particularly if the affected area is quite large and the objects are relatively small and scattered as it is for private houses in the residential area. For example, Californian wildfires past year caused significant damages in Ventura and Santa Barbara counties. The fires destroyed at least 1,063 and 5,643 structures in these areas respectively \cite{top20}.
The significant delay in time of Emergency Mapping also might be caused by the availability of remote sensing data which has it's physical limitations (cloudiness, day time, resolution etc.) as well as commercial ones (terms of use, costs etc.)
Needless to say that in the post-disaster recovery strategy the time is the key factor. That’s why we consider apply the Machine Learning and Computer Vision approach to the processing of Satellite and Aerial imagery to detect main damages and reduce the time costs.

%
%
%

\section{Existing approaches and solutions}
\begin{figure}
	\centering
    \includegraphics[width=0.8\textwidth]{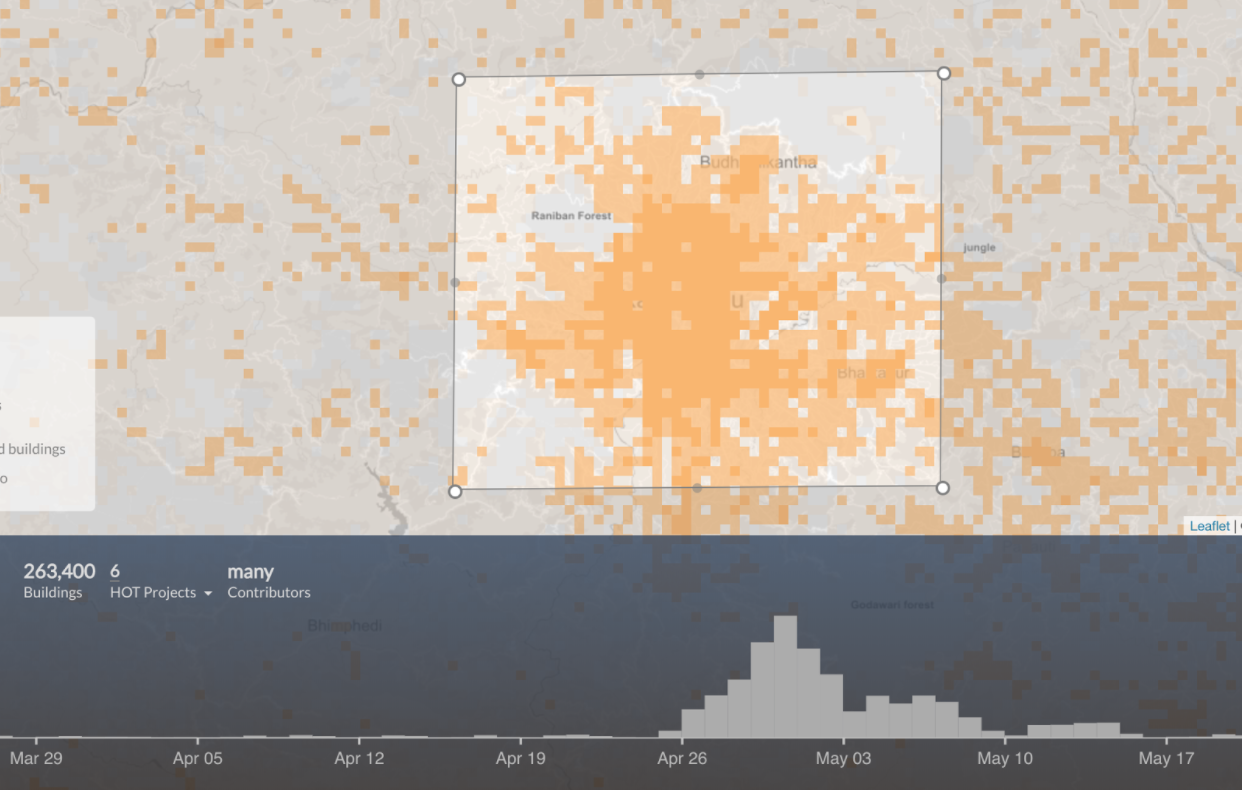}
    \caption{Time distribution of the building features created by OSM users for Kathmandu region (source - osm-analytics.org)}
	\label{nepal hot}
\end{figure}

Among the existing solutions for Emergency Mapping of disaster areas it’s worth to mention HOT that allows mapping of the chosen area in a collaborative way. Taking data for different areas, where HOT campaigns were activated, we estimated that the mapping process even in the areas of emergency that attract a lot of public attention, like Nepal earthquakes, takes several months to achieve the whole coverage (see figure \ref{nepal hot}).

Following the news of this incident we found several related media publications that provide assessments of damages. We assume that the work was doing manually on satellite images - comparing to the date of incident (Dec. 4) it's taken about six days to prepare maps for the article (Dec. 11 LA Times article update) \cite{lag1}. That most probably is caused by the amount of work needed to find appropriate data sources and make a damage map.

UNOSAT Rapid mapping service \cite{unosat} is a framework of United Nations Institute for Training and Research in the field of emergency mapping. Even though it’s quite challenging to estimate the time efficiency by their results since there is no clear definition of what states for ``rapid''. Usually the temporary lag of UNOSAT maps is two days from the satellite imagery acquisition date. 

One of the keys to the solution of the problem might be a deep learning approach. In the last few years the deep convolutional networks became a robust solution for many problems concerning image analysis. Different variants of the networks are able to solve the problems of semantic segmentation, object classification and detection  \cite{unet15, fcn, alexnet, maskrcnn}. 

The main drawback of this class of methods is that the deep convolutional networks need a big amount of training (previously manually marked ground truth) data. On the one hand, we can pre-train the method using the data about the other event of the kind, that took place in the past. But the results of this kind of training may be unpredictable due to the difference between the data in the training and test cases. These issues are concerned in our workflow that is proposed in the following section.

\section{Problem statement and proposed workflow}

The main problem we want to deal with is to decrease the time needed to retrieve crucial information for decision making in emergency situations when the proper remote sensing data is available. 
We propose the following workflow:
\begin{enumerate}
\item Determine the case of interest.
The deep learning methods work significantly better when the objects of interest are similar to each other, so the case should be narrow, for example “burned houses” or “flooded roads”.
\item Create a training dataset. 
The deep learning methods need a training data so they could learn the objects of interest and their properties. The training dataset consist of the real data (in our case, two aerospace images, one taken before the catastrophic event, and the other - after the event) and the labels that annotate and outline every damaged structure of the type.
\item Train and validate a deep learning method using the dataset. 
The method (or model) extracts the information from the training dataset. Its ability to detect the damaged objects of interest is validated using a subset of the training data. This pre-trained model will be used in every case of the forthcoming emergency of the given type. 
\item Obtain information of a new emergency case.
This is where our method starts working. The data should be of the same or similar type (spatial resolution, target objects, color depth) as that used for training, this is a critical requirement for the model to work properly.
\item Fine-tune the model for the new case.
Despite the similarity of the data, the model may be unable to process them correctly due to small differences, for example different sunlight. The fine-tuning can be done using automatically annotated data from the new case, or using the manual markup for a small subset of the data.
\item Run the automatic image processing.
Now that the model is trained for the case, we can make an automated processing of the rest of the data and have them ready for the decision making.
\end{enumerate}
Using this approach, we need to spend some time for creation of the reference training dataset, but normally it should be made before the emergency event. Then, after the event, the amount of work needed is much less that allows us to propose a fast working and thus efficient solution.

\section{The experiment}
To validate and demonstrate the workflow, we have chosen the case of wildfires in two areas of Ventura and Santa Rosa counties, California, USA, where many houses were burned to the ground in 2017. The advantage of this choice is justified by the availability of hi-resolution data provided by Digitalglobe within their Open Data Program \cite{opendata}.
In the following section we will follow our workflow on the case, and describe both general principles of the deep learning application to the imagery and our steps in the particular case.
\subsection{Determine the case of interest}
\begin {figure}
\subfigure[Before]{\includegraphics[width=60mm]{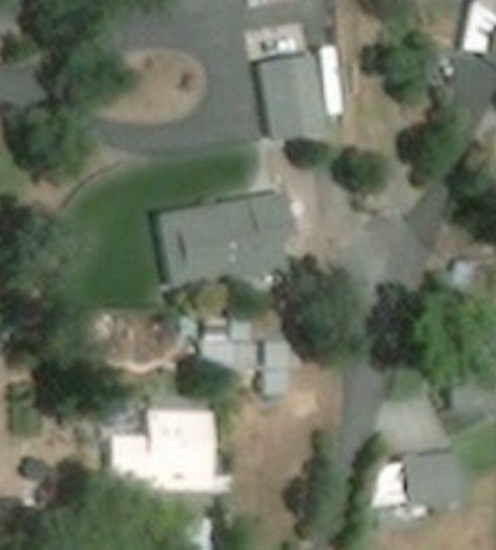}}
\subfigure[After]{\includegraphics[width=60mm]{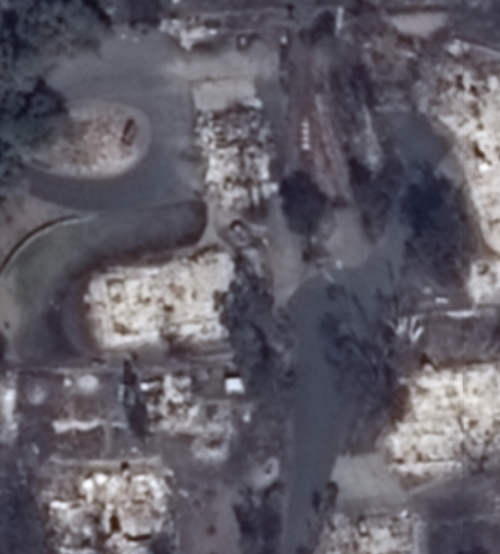}}
\label{case_interest}
\caption{Satellite images of buildings before and after the fire event}
\end{figure}

In our research the case of interest is “houses destroyed by fire”. A typical example of the object of interest is depicted in figure \ref{case_interest}. 

It is worth noting, that the case should be restricted as narrow as it is possible for it makes a big difference when speaking about the deep learning methods. For example, if we train the method on the images like this, where the houses are completely destroyed, it will not be able to detect partially damaged buildings. Also the type of building and even the rooftop material can change the result significantly.

\subsection{Create a training dataset}
\begin {figure}
\subfigure[map]{\includegraphics[width=0.5\textwidth]{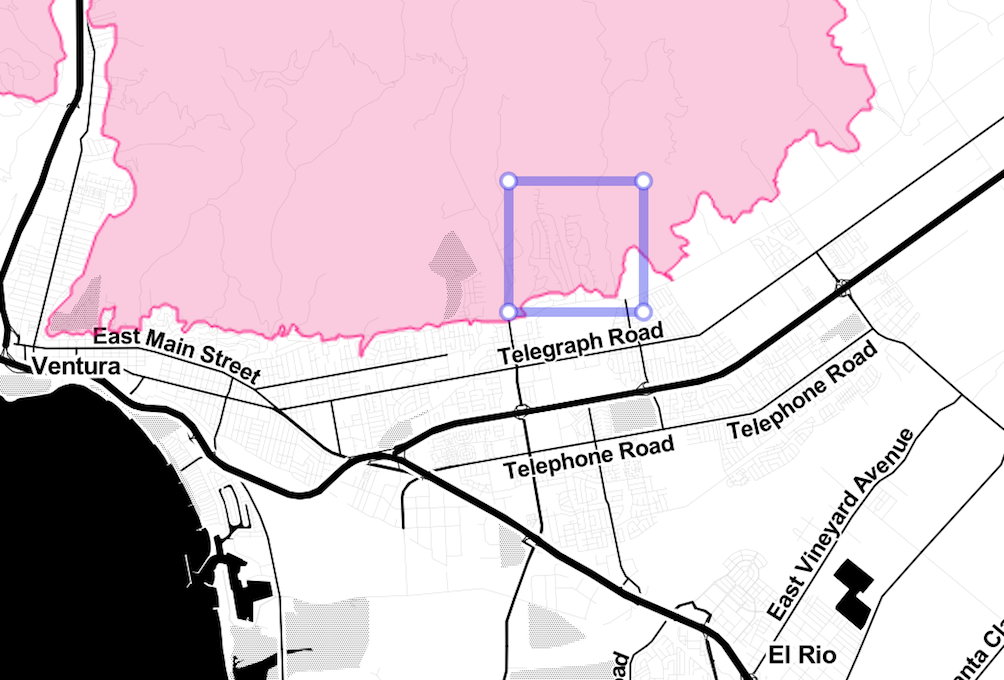}}
\subfigure[markup]{\includegraphics[width=0.4\textwidth]{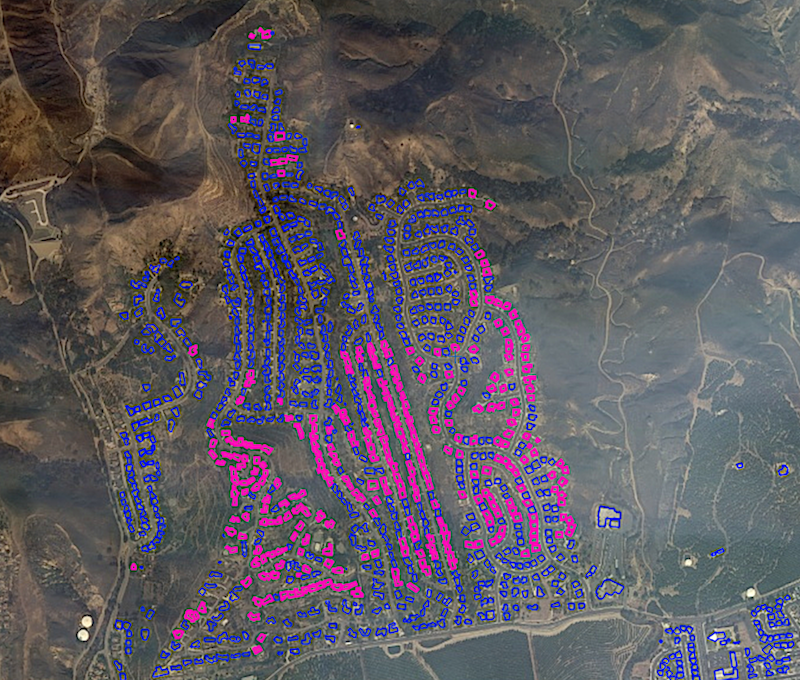}}
\label{thomas_fire}
\caption{Training area in Ventura and the resulting markup (Openstreetmap, Stamen design; Digitalglobe. Bounding box coordinates:
-119.2278	34.2836	
-119.1916	34.3065)}
\end{figure}

The training area is chosen in the Ventura, Santa Barbara, California, that was severely affected by the Thomas Fire in the December, 2017 (see figure \ref{thomas_fire}).

\begin {figure}
\centering
\includegraphics[width=60mm]{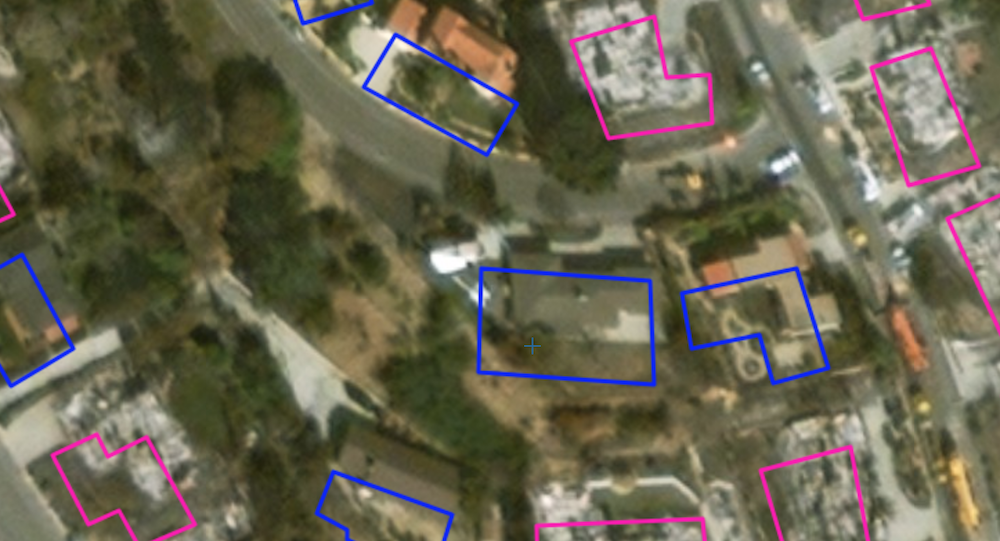}
\label{figure_misalign}
\caption{Misalignment of the Digitalglobe image with OSM markup}
\end{figure}

The preferable source for high-resolution data is the Digitalglobe Open Data Program. This program is one of the few sources of free high-resolution data, and the data is distributed early after the event \cite{opendata}. However, in the case of South California the existing Openstreetmap (OSM) mapping which was used as the initial input for the markup is based on Google Maps / Bing Maps imagery that is better orthorectified, so that the image coordinates differ, as it can be seen in the figure \ref{figure_misalign}. This makes existing map not as good source of the ground truth.

Due to these reasons, we had to use the Google Maps imagery, that is similar to the Digitalglobe open data in terms of image characteristics (similar spatial resolution, also 3-band 8-bit per band images), but both pre-event and post-event images are available, and the better orthorectification leads to good alignment with the OSM.

The crowdsourced mapping data from OSM were not full and did not contain the information about burned buildings, so it was necessary to improve the OSM markup before using it as the ground truth for training the deep convolutional network. We facilitated the manual work by using of OSM as the source of initial vector features, selecting all the ones tagged as “building”. All the extracted features than were checked through the visual inspection and annotated with the appropriate additional tag ``disaster''=``damaged\_area'' if the one was destroyed by the fire.
To complete training dataset we used cartographic tools as Openstreetmap ID which is open source editor for online mapping for OSM \cite{osm_id}. The final markup contains 760 not damaged buildings and 320 ruined buildings (see figure \ref{thomas_fire}) was exported in GeoJSON format using OSM API and additionally processed using our Python stack tool to convert and rasterize vector data into 1-band pixel masks. 

\subsection{Train and validate a deep learning method using the dataset}
We used a semantic segmentation approach to the change detection. The semantic segmentation of an image results in a mask of the pixels that are considered to be of the target class or classes. 
In our case, when we have two images - before and after the event - we can gain maximum from the given data if we stack them together and make a 6-band image (3 bands before and 3 bands after). A convolutional network for change detection was built in the encoder-decoder manner, which has great success in solving semantic segmentation problems \cite{unet15}. For a model that works with pairs of 3-band images, one could use a single 6-channel encoder, but this would not allow the use of a transfer-learning technique to speed up learning and improve the final quality of the results, so the model was built on a two-stream encoder, each of which “looked at its own” 3-band image and one common decoder. This approach made it possible to use the pre-trained on ``ImageNet'' classification dataset \cite{keras} weights for the RGB images independently in each of the branches of the encoder.

\begin {figure}
\includegraphics[width=\textwidth]{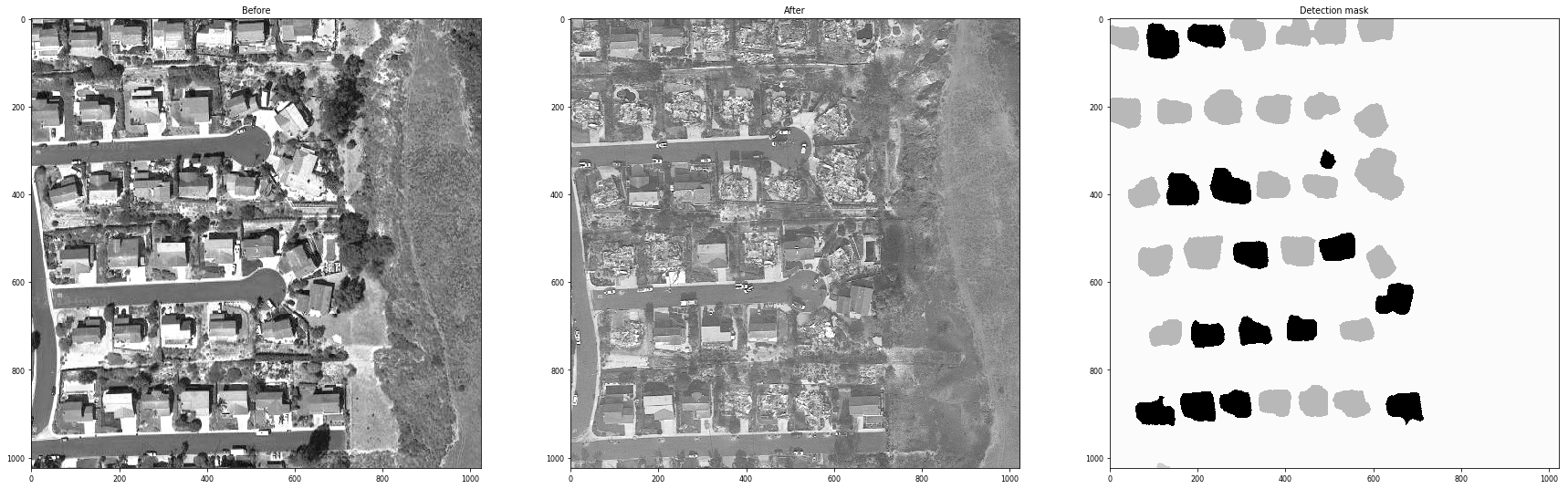}

\label{figure_result1}
\caption{Results of the change detection on the validation subset of data in Ventura. Left: image taken before fires, center: image taken after fire, right: segmentation results black - non-damaged, gray - damaged buildings}
\end{figure}

Validation on the part of the Ventura dataset that was not used for training gave appropriate results, see figure \ref{figure_result1}. Pixel-wise $F1$-score for the class of burned buildings is $0.859$ and for the class of unburned buildings is $0.818$.

Overall training of the model has taken less than an hour  on a Tesla P100 GPU using Keras with Tensorflow backend \cite{keras, tensorflow}.
The trained model is now ready for the new cases of the massive fire.

\subsection{Obtain information of a new emergency case}
\begin {figure}
\centering
\includegraphics[width=.6\textwidth]{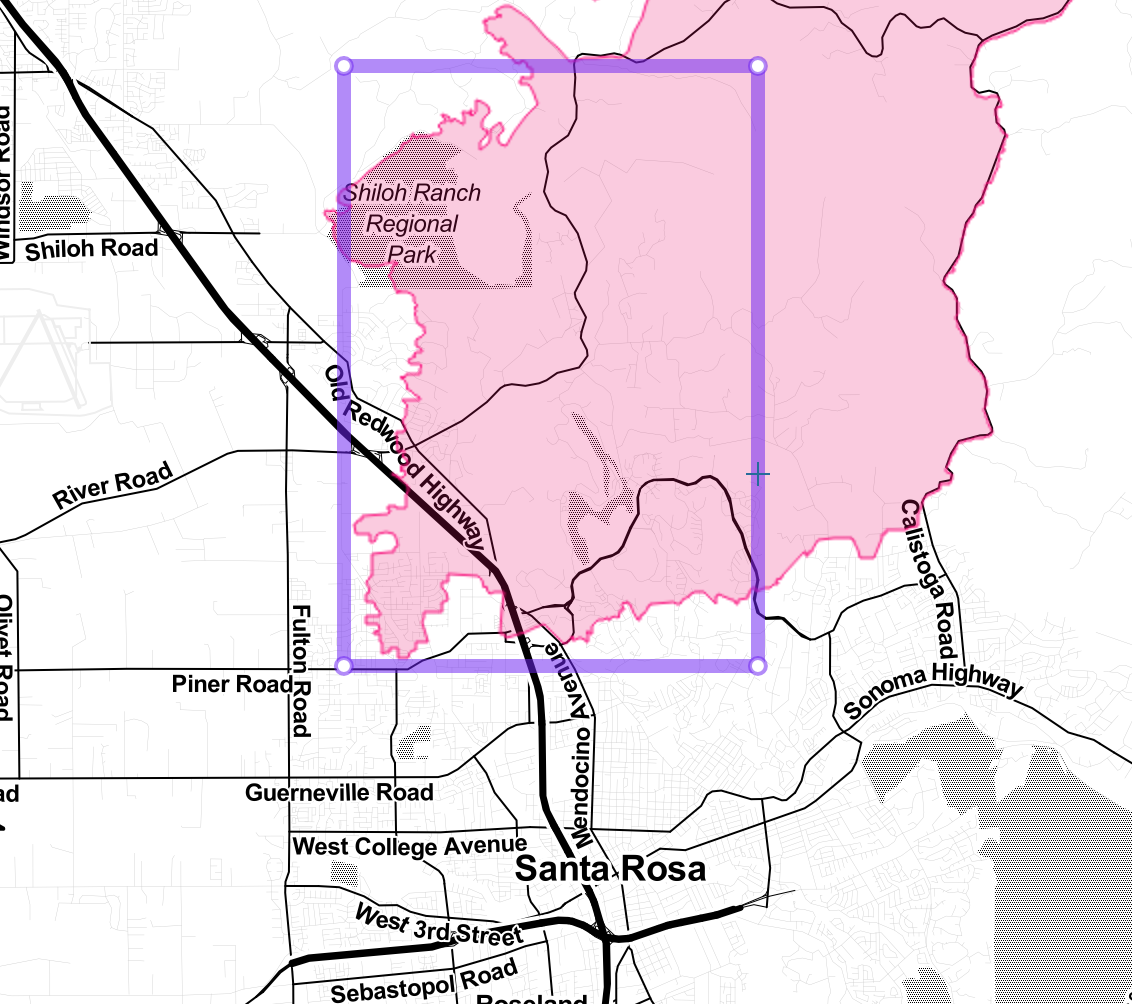}
\label{figure_tubbs}
\caption{A map of the new zone of the Tubbs fire in Santa Rosa, California}
\end{figure}

We consider the fire in Santa Rosa, California (Tubbs Fire, October 2017) as the ``new case'' of the same type of events (see figure \ref{figure_tubbs}).  The Open Data program has images both before and after the fire event, so we can use them for the test.

\begin {figure}
\includegraphics[width=\textwidth]{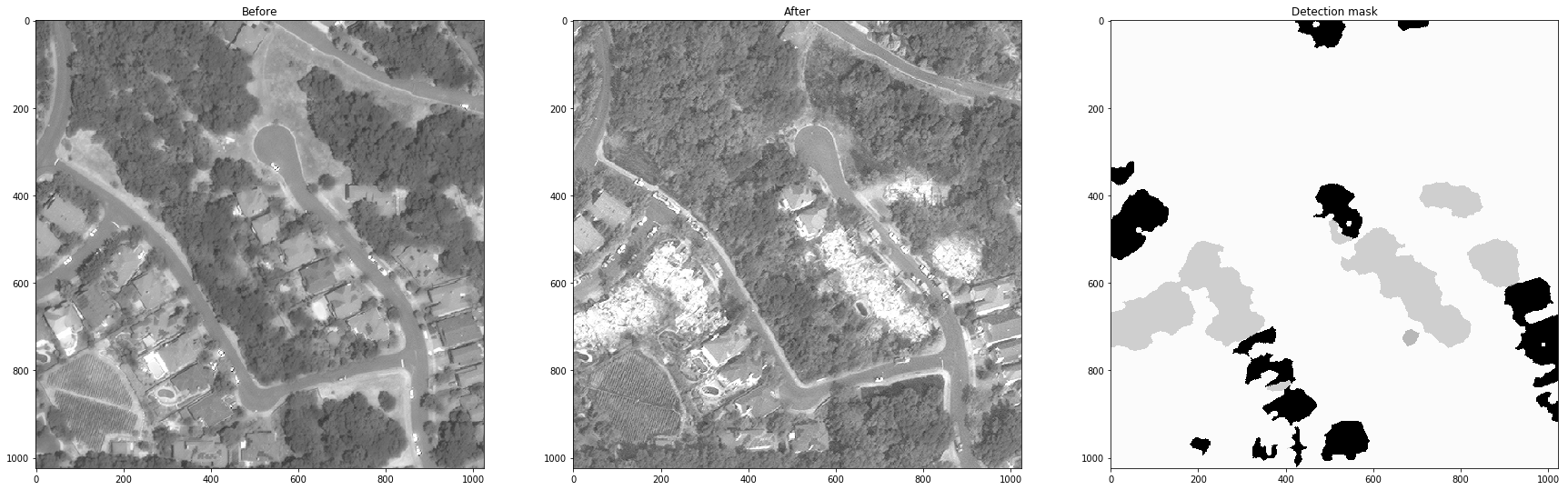}
\label{figure_result2}
\caption{Results of the change detection on the test subset of data in Santa Rosa without fine-tuning. Left: image taken before fires, center: image taken after fire, right: segmentation results black - non-damaged, gray - damaged buildings}
\end{figure}

As the data in this case have similar characteristics, we tried the image segmentation with the model as is, without any changes. The result is unsatisfactory, however it does have some correlation with the real data. This can be caused by differences in season, solar angle, image preprocessing difference, or by some difference in she structure of the residential areas themselves. For example, buildings in Santa Rosa are closer to each other. The example of the results see in \ref{figure_result2} 

\subsection{Fine-tune the model for the new case}

\begin {figure}
\centering
\includegraphics[width=0.8\textwidth]{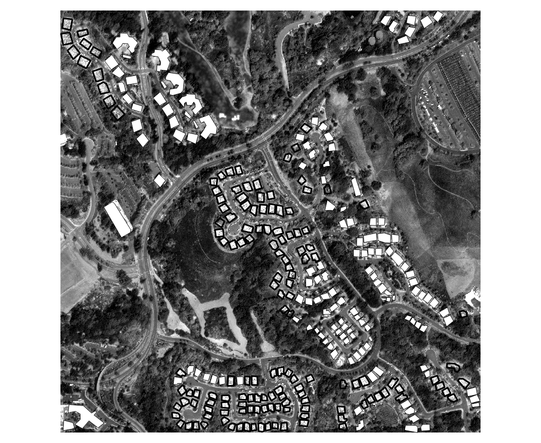}
\label{figure_markup2}
\caption{Small dataset for fine-tuning of the net to the new data. White - unburned buildings, black contours - burned buildings}
\end{figure}

The results above show that we need to train the model for the new area. In order to do this, we make a new small dataset in a part of the Santa Rosa, see figure \ref{figure_markup2}. It contains 146 burned and 137 undamaged houses, so it requires far less time and effort. The preparation of the dataset took about an hour of manual markup by one person.

Switching from one part of dataset to another, the results of the model were greatly deteriorated, but the dense marking of just less than 300 houses on new images allowed to improve the quality on the whole new data and reach almost the same result for 10 minutes of additional training.

\subsection{Run the automatic image processing}
\begin {figure}
\subfigure[Before event]{\includegraphics[width=0.5\textwidth]{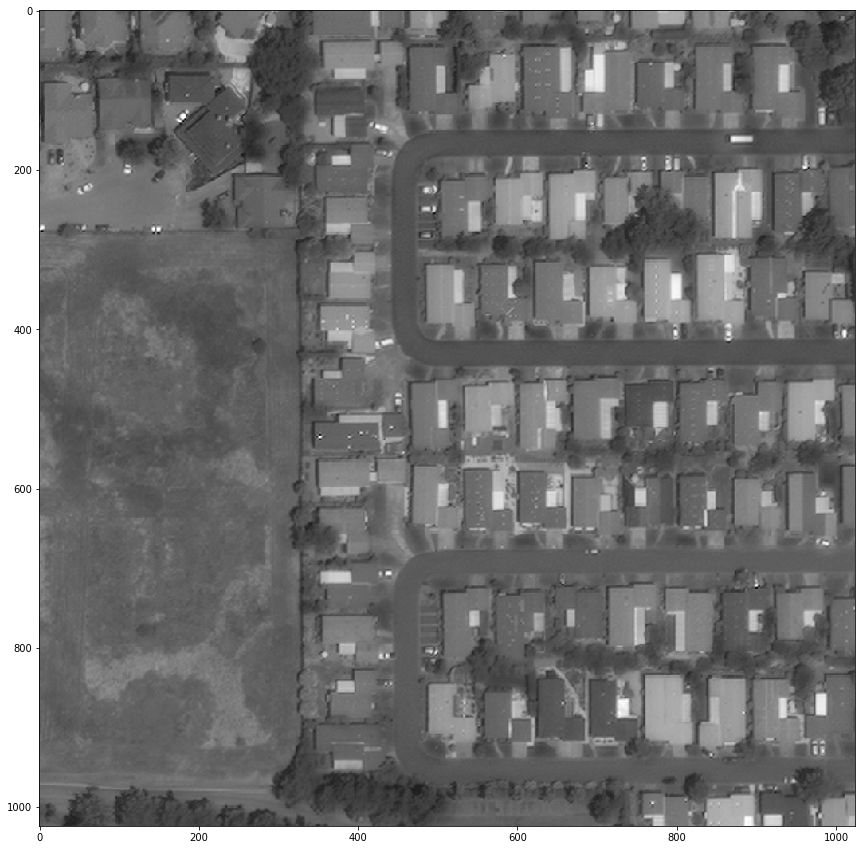}}
\subfigure[After event]{\includegraphics[width=0.5\textwidth]{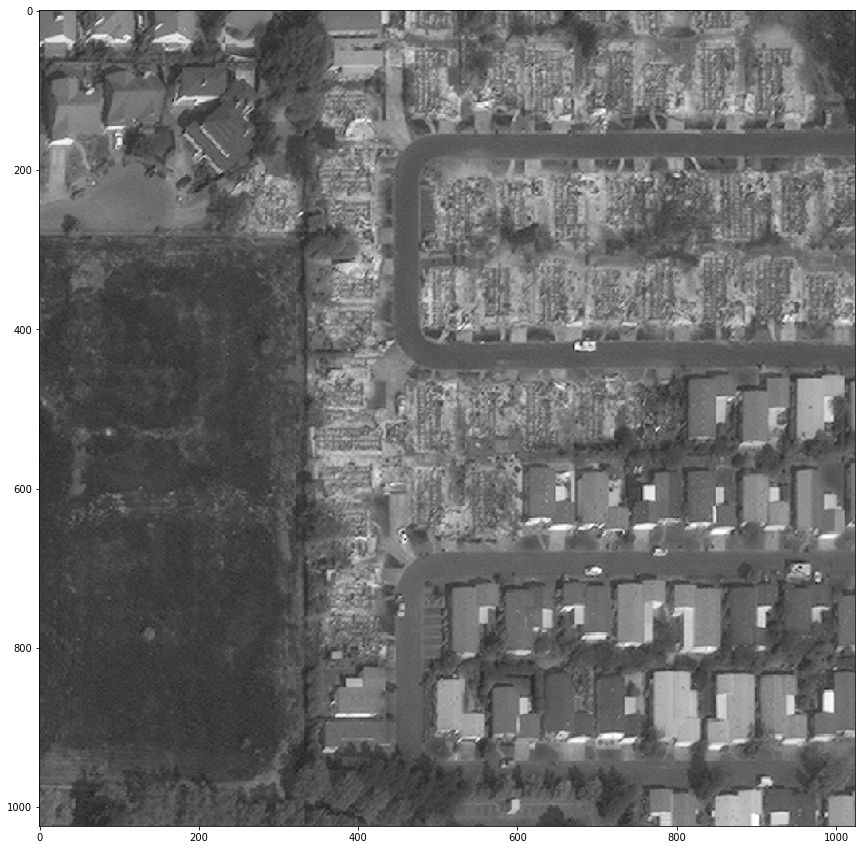}}
\label{figure_example3}
\caption{An example of the test area image before and after the fire}
\end{figure}

\begin {figure}
\subfigure[No fine-tuning]{\includegraphics[width=0.5\textwidth]{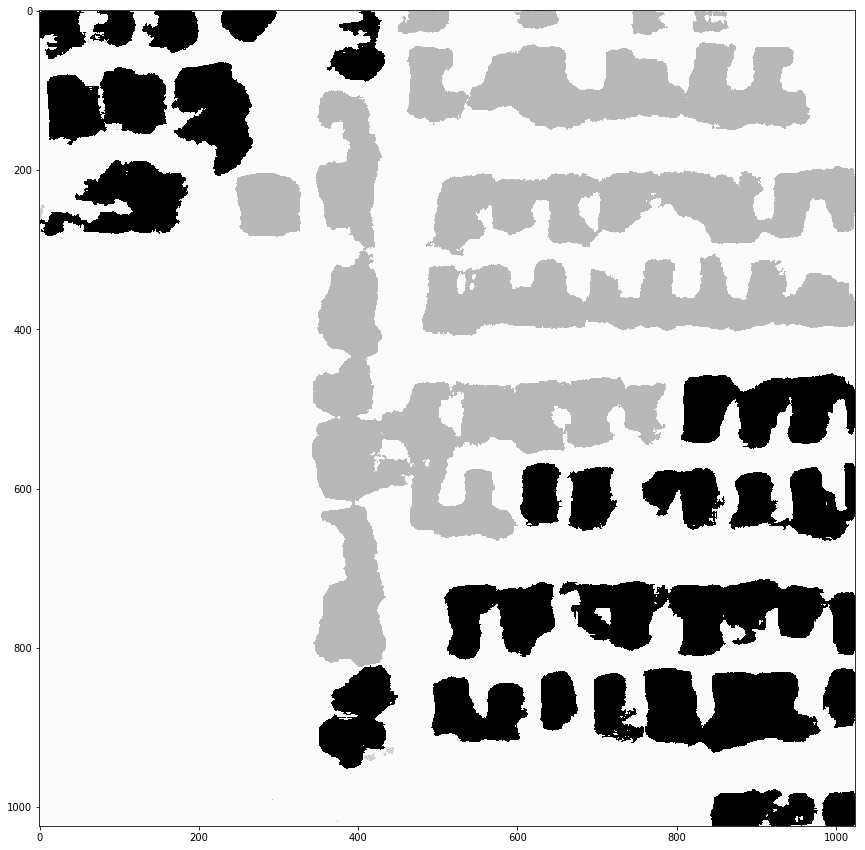}}
\subfigure[With fine-tuning]{\includegraphics[width=0.5\textwidth]{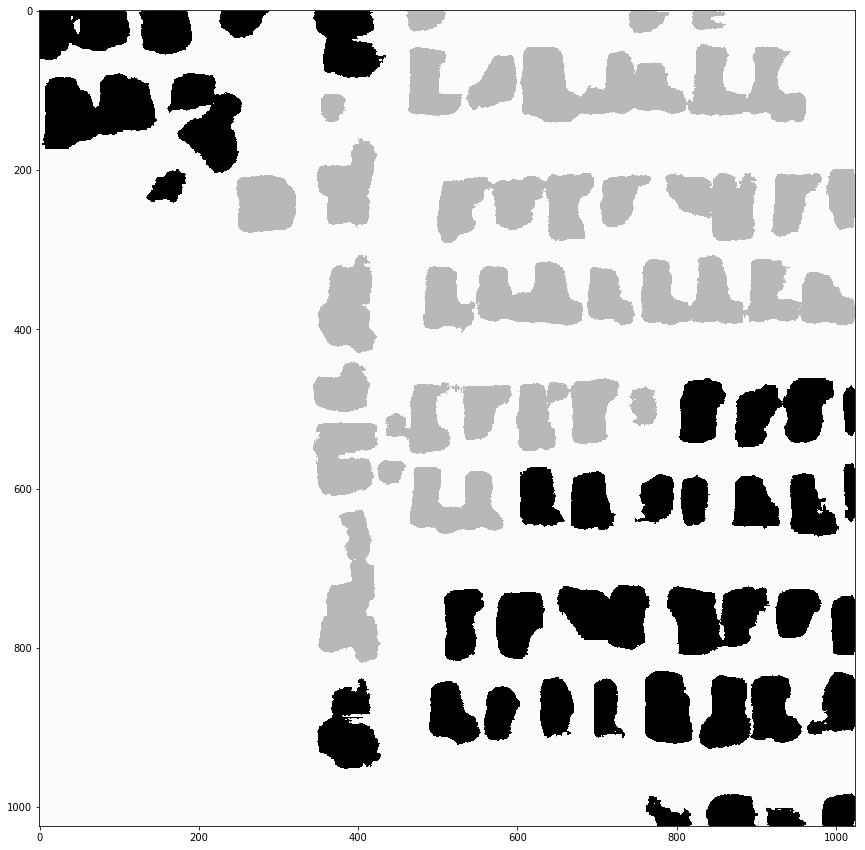}}
\label{figure_result3}
\caption{Results of the image segmentation before and after fine-tuning}
\end{figure}

The rest of the Santa Rosa region of interest was processed automatically by the trained model.
The example of the result taken from the test zone in the center of Santa Rosa town is shown in figures \ref{figure_example3}, \ref{figure_result3}. It can be clearly seen that non fine-tuned method tends to merge the regions of the separate buildings into one area, while after the fine-tuning the resulting regions can be easily separated at the post-processing stage. 

After the fine-tuning, the change detection method can give very good results on image segmentation, and even give a good chance to distinguish between separate houses that is very important in the task of damage assessment when it is necessary to estimate the accurate number of damaged structures. 

Note that the segmentation approach is more robust than the detection one because it allows to estimate the area of the target changes, that can be necessary in other emergency cases like floods, blockages, fire damage to crops or forests etc.

\section {Time efficiency}

The manual markup of our Ventura training area (figure \ref{thomas_fire} ) should take about $1.5 - 2$ days by a qualified specialist, assuming that mapping of buildings features takes averagely 30 sec. per feature. But more realistic is the time evaluation of HOT mapping as it represents the real rate of community. Besides, the HOT tools are built on the top of OSM Data Base and are planned not for mapping of the state of the objects like “burnt buildings” but to improve the basic maps when the cartographic data is missed and needed by humanitarian and emergency organizations. 

The full workflow for the new area, where we had to make the only a small training subset took about 3 hours including model training and automatic processing.

That gives us less time needed for information retrieval for the emergency management.

\section{Further research}
At the current stage we have developed a workflow and a method of the damaged areas segmentation. In the further research we will continue the development of the segmentation method to increase its accuracy and robustness to the data characteristics changes.

The method can be also extended to the problem of instance segmentation, that is distinguishing between separate objects, counting the objects, and converting them to the map that can be used online and in GIS applications.

We will apply the approach to the Open Data in the case of new events of this domain, the other types of disasters such as floods and tsunami, and will extend the training dataset to extrapolate this approach to the other cases and territories.

\section{Conclusion}
We’ve formulated the problem based on the research of the tools and frameworks for disaster mapping. Based on the problem, we proposed a workflow involving deep learning and use of open data for the emergency mapping.

We’ve created the training and test datasets for California fires, which means the raster mask of the vector features of damaged and non damaged buildings in the area and the appropriate pre- and post-event imagery to develop a change detection method and validate the approach.

We developed a method of change detection, based on convolutional neural networks, that is able to make semantic segmentation of the area subjected to massive fires, mapping burned buildings. The method can be easily extended to the new areas and data sources with a little training for the data peculiarities (fine tuning).

The workflow turned up to give substantial profit in terms of time needed for emergency mapping and in the future we will extend it to the other cases.

%
%

%
\end{document}